\documentclass[10pt,twocolumn,letterpaper]{article}
\usepackage{multirow}
\usepackage{multicol}

\usepackage{wacv}              
\usepackage[accsupp]{axessibility}

\usepackage{graphicx}
\usepackage{amsmath}
\usepackage{amssymb}
\usepackage{booktabs}
\usepackage{adjustbox}

\usepackage{pifont}%
\usepackage[linesnumbered,ruled,vlined,english,onelanguage]{algorithm2e}

\usepackage[pagebackref,breaklinks,colorlinks]{hyperref}

\usepackage[capitalize]{cleveref}
\crefname{section}{Sec.}{Secs.}
\Crefname{section}{Section}{Sections}
\Crefname{table}{Table}{Tables}
\crefname{table}{Tab.}{Tabs.}

\def\wacvPaperID{41} 
\def\confName{WACV}
\def\confYear{2024}
\newcommand{\cmark}{\ding{51}}%
\newcommand{\xmark}{\ding{55}}%
\begin{document}

\title{TextAug: Test time Text Augmentation for Multimodal Person Re-identification}

\author{
Mulham Fawakherji\textsuperscript{1},
    Eduard Vazquez\textsuperscript{1},
    Pasquale Giampa\textsuperscript{1}, Binod Bhattarai\textsuperscript{2}
    \\[1em]
    \textsuperscript{1}Fogsphere, London, UK\\
    {\tt\small \{Mulham.Fawakherji, Eduard.Vazquez, Pasquale.Giampa\}@fogsphere.com}
        \and
    \textsuperscript{2} University of Aberdeen, Aberdeen, UK\\
    {\tt\small binod.bhattarai@abdn.ac.uk}
}
\maketitle

\begin{abstract}
Multimodal Person Re-identification is gaining popularity in the research community due to its effectiveness compared to counter-part unimodal frameworks. However, the bottleneck for multimodal deep learning is the need for a large volume of multimodal training examples. Data augmentation techniques such as cropping, flipping, rotation, etc. are often employed in the image domain to improve the generalization of deep learning models. Augmenting in other modalities than images, such as text, is challenging and requires significant computational resources and external data sources.
In this study, we investigate the effectiveness of two computer vision data augmentation techniques: ”cutout” and ”cutmix”, for text augmentation in multi-modal person re-identification. 
Our approach merges these two augmentation strategies into one strategy called "CutMixOut" which involves randomly removing words or sub-phrases from a sentence (Cutout) and blending parts of two or more sentences to create diverse examples (CutMix) with a certain probability assigned to each operation. This augmentation was implemented at inference time without any prior training. Our results demonstrate that the proposed technique is simple and effective in improving the performance on multiple multimodal person re-identification benchmarks.
\end{abstract}

\section{Introduction}
\label{sec:intro}
Person re-identification (ReID) is an important and popular research problem in the computer vision community~\cite{survey}. Existing data-driven methods mostly rely on the image domain~\cite{Zhou_2019_ICCV}.
Recently, the use of text queries for person re-identification has gained significant attention, particularly in the form of search queries in natural language descriptions \cite{Chen_2018,8100034,Li2017IdentityAwareTM}. These models enable the retrieval of individuals in images using verbal or written text descriptions, making them appealing to human nature. However, this approach presents additional challenges in computational modeling, including the high cost of obtaining accurate and comprehensive training data and the difficulty of consistently and reliably modeling, as well as interpreting it for arbitrary images. Poor-quality surveillance images and limited access to an appropriate dataset further complicate this.

In recent years, there has been significant research interest in embedding representations of images and text into a joint space for various tasks. This approach has been successfully applied to tasks such as image annotation and search \cite{7299073,8100140}, zero-shot recognition \cite{NIPS2013_2d6cc4b2,7780382}, robust image classification \cite{NIPS2013_7cce53cf}, image description generation \cite{https://doi.org/10.48550/arxiv.1412.2306}, visual question-answering \cite{Nam2016DualAN}, and more.
Overall, the ability to embed representations of images and text in a joint space offers numerous potential benefits for a wide range of tasks.

In addition to utilizing multi-modal input, data augmentation has been widely used in computer vision (CV)
over the past decade. Geometric augmentations on images such as cropping, reflection, and translation \cite{krizhevsky2017imagenet} have become standard techniques in CV and have rapidly become a common way to improve the generalization of computer vision models \cite{Simonyan2015,He2016}. In particular, these techniques have been used to increase the size of the training dataset and mitigate overfitting.
However, augmenting data in natural language processing (NLP) has been a challenging task. A simple approach like synonym replacement \cite{Zhang2015} \cite{Wieting17} requires an external data source (such as a thesaurus) and introduces additional engineering costs for the synset selection algorithm. Other methods like pre-trained language models \cite{Kobayashi2018} or translation systems \cite{Edunov2018,Wieting17} are required to rephrase the training text, limiting the set of languages that can benefit from data augmentation and making low-resource languages even more challenging compared to English. Furthermore, these methods require significant computational resources.
To address these challenges, NLP researchers have proposed the application of computer vision methods like Mixup \cite{Guo2019AugmentingDW,Jindal2020,Guo_2020}
which combines pairs of examples by taking weighted linear combinations of their input data and labels. This approach generates new synthetic examples that are a linear interpolation of the input examples, thereby increasing the diversity of the training data and improving the model's generalization ability. Mixup is a soft augmentation technique that smoothly blends the input examples to generate new examples, and the resulting examples often retain the features of both input examples.

Cutout~\cite{cutout} and CutMix~\cite{yun2019cutmix} are popular data augmentation techniques used in computer vision to enhance the robustness and generalization of deep learning models. 
Unlike Mixup, our proposed augmentation CutMix and Cutout are hard augmentation techniques that manipulate the input data by removing or replacing parts of the input examples and introduce a higher degree of variation to the input examples, often leading to more significant changes in the input data. This can enhance the model’s resilience to noise and distortions in the input data in NLP. These techniques are effective in increasing the size of the training data set and preventing overfitting.
In our study, we considered having only one sentence description for each person, which makes it difficult to implement Mixup algorithms that require pairs of examples. Therefore, we focused on utilizing CutMix and Cutout techniques since they can be applied to single examples and still introduce variation in the data.

\begin{figure}[]
    \centering
    \includegraphics[width=0.45\textwidth]{ 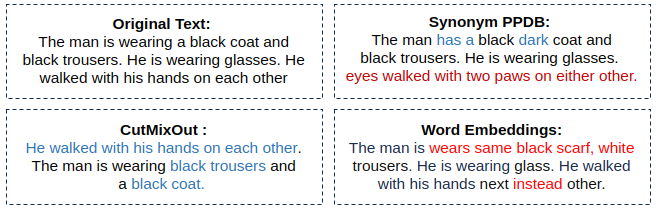}
  \caption{Displays a comparison of different text augmentation strategies used as input for person re-identification models. The first row, from left to right, shows the original text and the text augmented using synonym replacement from the PPDB dataset. The second row, from left to right, shows the text augmented using Cutout and the text embeddings generated using Word2Vec. }
    \label{fig:text_aug}
\end{figure}

\begin{figure*}[]
    \centering
    \includegraphics[width=0.7\textwidth]{ 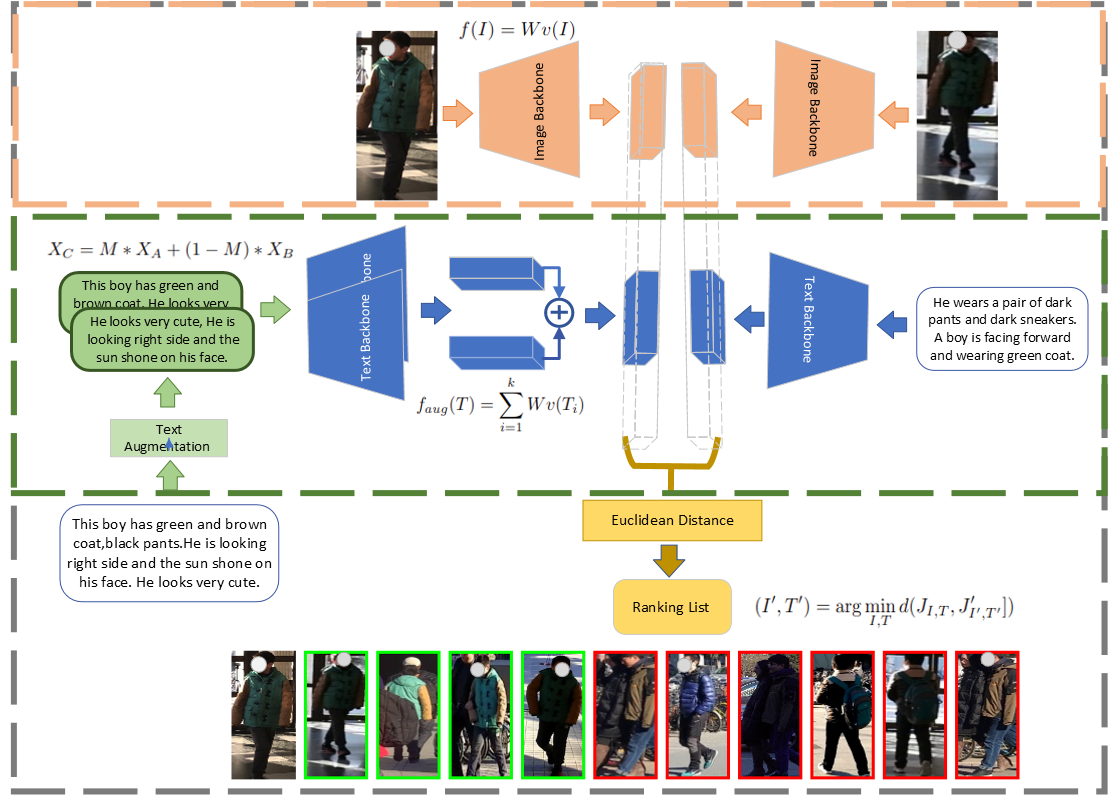}
    \caption{The figure illustrates our architecture for person re-identification, which utilizes both text and image inputs. During inference, the text input is augmented to generate multiple representations and fed into a Clip pre-trained text backbone model. The resulting embeddings are summed to create a single text embedding. The image input is also fed into a Clip pre-trained image backbone model. The text and image embeddings are concatenated to generate a single embedding vector, which is used for re-identification by measuring the distance between the query and gallery vectors. }
    \label{fig:mesh1}
\end{figure*}

We adapted these methods for augmenting text descriptions and employed them in the context of multimodal person re-identification. Figure~\ref{fig:text_aug}
compares our approach with some of the important existing text-augmentation methods. Unlike Synonym PPDB~\cite{PPDB} and Word Embedding~\cite{wang-yang-2015}, our approach is simple and does not require a Thesaurus or external database such as WordNet.
Synonym PPDB, which substitutes some of the original words with their synonyms, has produced a sentence that is grammatically correct but lacks coherence and accuracy in describing the original scene. The phrase "eyes walked with two paws on either other" seems to be a result of an error in the synonym substitution process and does not make sense in the context of the scene.  Similarly,  Word Embeddings, which generates variations of the text by substituting words with similar meaning, has produced a sentence that is similar to the original text, but with slight differences in wording. However, the technique seems to have missed some important details of the scene, such as the fact that he is wearing a black coat and black trousers. On the other hand, Cutout and Cutmix have produced variations of the original text that are coherent and precise in describing the scene. Cutout has masked out a random portion of the image, while Cutmix has blended two different images by cutting and pasting a random portion of one image onto another. Both techniques have retained the main elements of the scene, such as the man's clothing while introducing minor variations in the phrasing.
Our method involves applying text augmentation techniques to the query input, which comprises text and images.
Figure~\ref{fig:mesh1} summarizes our pipeline. The text is processed to generate multiple representations using text augmentation methods.
These representations are then fed to a text encoder, which generates corresponding text embeddings. Similarly, the image is passed through an image encoder, which generates image embeddings. The generated text embeddings are aggregated together, and the resulting aggregation is concatenated with the image embedding to form the final representation of the query input.
In our pipeline, we utilized a pre-trained CLIP model\cite{Clip}, which is trained on large amounts of diverse text and image data and can generate high-quality text and image embeddings that capture the semantic and visual information in the input data.

To sum up, one of the key contributions is a simple yet effective text augmentation technique called $CutMixOut$ inspired by cutout~\cite{DeVries2017} and cutmix~\cite{Yun_2019}. We employ it during inference for multimodal person re-identification. To the best of our knowledge, this
is the first work employing text augmentation in multimodal person re-identification. From the extensive experiments, we found it highly effective for person re-identification. Our results demonstrate that the proposed approach can significantly enhance the generalization of a model without further training.

\section{Related Work}
\label{sec:rel}
Previous research on ReID has primarily focused on developing discriminative features such as foreground histograms \cite{Chowdhury}, bag-of-visual words \cite{Zhang2015}, or hierarchical Gaussian descriptors\cite{Matsukawa2016}. However, ReID can also be approached as a metric learning problem, which requires a reasonable distance measurement for inter- and intra-class samples \cite{Koestinger2012}. These two approaches have been combined in deep neural networks \cite{yi2014}, where the parameters are optimized using an appropriate loss function. The use of CNNs has become common, with ResNet-50 \cite{He2016} being a popular model for most ReID datasets. However, CNNs have been criticized for highlighting irrelevant regions due to overfitting limited training data. To address this, lightweight models such as OSNet \cite{Zhou2019} have been developed, and network architecture search has been used to create compact models. Additionally,  a researcher \cite{Zhang2021} has proposed a data selection method to choose generalizable data during training. While these methods have shown good results on small datasets, their performance drops significantly on larger datasets such as MSMT17 \cite{wei2018person}, mainly due to overfitting.
Incorporating prior knowledge into a neural network can help prevent overfitting. One approach is to use features from different regions for identification, which can be achieved by dividing the feature into horizontal stripes. The PCB \cite{Sun2018} and SAN \cite{Qian_2020} models enhance the network's ability to represent the local region in this way. The MGN \cite{Wang_2018} model takes this approach further by utilizing a multiple granularity scheme on feature division and has several branches to capture features from different parts. However, this results in increased model complexity. The BDB \cite{Dai2019} model has a simpler structure with only two branches, one for global features and the other for local features, and employs a batch feature drop strategy to randomly erase a horizontal stripe for all samples within a batch. CBDB-Net \cite{Tan_2021} builds on this by enhancing the feature-dropping strategy.

Recently the use of text queries for personal search has gained significant attention, particularly in the form of search queries in natural language descriptions \cite{Chen_2018,8100034,Li2017IdentityAwareTM}. 
These models allow for performing image searches using natural language, making it more natural and understandable. This is a key factor in successfully increasing the general reach of a method in the real world.
However, this approach presents additional challenges in computational modeling, such as the high cost of obtaining accurate and rich training data and the difficulty of consistently and reliably modeling rich and complex sentence syntax, as well as interpreting it for arbitrary images. This is further complicated by poor-quality surveillance images and limited access to a proper dataset.
In recent years, there has been significant research interest in embedding representations of both images and text into a joint space for a variety of tasks. This approach has been successfully applied to tasks such as image annotation and search \cite{7299073,8100140}, zero-shot recognition \cite{NIPS2013_2d6cc4b2,7780382}, robust image classification \cite{NIPS2013_7cce53cf}, image description generation \cite{https://doi.org/10.48550/arxiv.1412.2306}, visual question-answering \cite{Nam2016DualAN}, and more.
Overall, the ability to embed representations of images and text in a joint space offers numerous potential benefits for a variety of tasks.

\section{Methodology}
\label{sec:meth}

\par Given a set of $N$ gallery images ${I_1, I_2, ..., I_N}$ and a query image $Q$, the goal of a person re-identification is to find the image in the gallery set that is most similar to the query image. 
The process of generating embeddings involves mapping each image to a d-dimensional vector that captures the essential features of the image. In the past, hand-crafted features~\cite{ma2012local}, distance metric learning~\cite{mignon2012pcca} and deep learning architectures~\cite{li2021diverse} have been employed to learn the features $ f(I_i) = \phi(I_i) $.
where $I_i$ is the $i$th image in the set, $\phi$ is the image model, and $f(I_i)$ is the embedding vector for image $I_i$.

Once the embedding vectors are computed, the distance between the query embedding and each image embedding can be measured using a distance function such as the Euclidean distance or cosine distance. The similarity score between the query image and each image in the set can be computed as:~$ S(I_i, Q) = d(f(I_i), f(Q)) $, where $Q$ is the query image, $f(Q)$ is the embedding vector for the query image, and $d$ is the distance function. Unlike most previous works that deal with images only, we are dealing with multi-modal data for a person re-identification.

Multimodal person re-identification involves identifying individuals across different sources of data, such as images, videos, and text. While visual features have traditionally been the main focus of person re-identification, text can also play an important role in this task. 
Text can provide valuable information about a person's identity, which can include a wide range of information related to a person's appearance, including the color and style of their clothing
This information can be used to supplement visual features, especially in cases where the visual information is incomplete or ambiguous.
A multimodal person re-identification involves learning the joint representation of data from multiple sources and comparing the query with the gallery similar to that of an image-based 
person re-identification. In this paper, we use textual descriptions along with images. 

Let $T_i$ be the textual description for image $I_i$ and $f(I_i)$ and $g(T_i)$ be their embedding representations, respectively. A multimodal person re-identification can be formulated as shown below.
$$E(I_i) = [f(I_i),g(T_i)]$$
$$E(Q) = [f(Q),g(Tq)]$$
$$ S(I_i, Q) = d(E(I_i), E(Q)) $$
In the above equations, $Q$ is Query, and $d$ is the distance function between the joint embedding vectors.
Our contribution lies in generating textual embedding by employing cut-out and mixing textual descriptions.

\subsection{TextAug: Text Augmentation}

Our proposed method for enhancing the performance and robustness of deep learning models in the task of person re-identification involves the use of inference time text augmentation techniques, specifically cutout and cutmix. These techniques, which are commonly used in computer vision tasks such as image classification, involve randomly removing a portion of an image or blending parts of multiple images.
We apply these techniques to the text portion of the query input to create multiple representations of the text. 
In CutMix, the binary mask M is used to select which sub-sentences from A should be used to create the new example $X_C$. The mask is created by randomly selecting a contiguous subsequence in A and replacing its sub-sentences with the corresponding sub-sentences from a copy of A. This process creates a new example that combines the sub-sentences from A.
The resulting example $T_{mix}$ is calculated by element-wise concatenating the sub-sentences from A and its copy according to the binary mask M. This creates a new example with a mix of sub-sentences from both parts of A, resulting in a more diverse dataset.
The mathematical representation for CutMix with a single sentence A is as follows:
Let $A = [a_1, a_2, ..., a_n]$ be a single sentence of length n substances.
Let $A'$ be a shuffled version of $A$, denoted as $A' = [a'_1, a'_2, ..., a'_n]$.
The resulting example $X_C$ can be calculated as:
\begin{equation}
    T_{mix} = M * X_{A} + (1 - M) * X_{A'}
\end{equation}
In CutOut, the binary mask M is used to remove parts of the input text A to generate the cutout text representation $X_C$. This process involves randomly selecting contiguous subsequences in A and replacing them with a special token that indicates they have been removed.
The resulting text representation $T_{out}$ is obtained by element-wise concatenating the remaining sub-sentences from A according to the binary mask M. This creates a new text representation with parts of the original input text removed, resulting in a more robust dataset for training NLP models.
The mathematical representation for CutOut with a single sentence A is as follows:
Let $A = [a_1, a_2, ..., a_n]$ be a single sentence of length n.
Let $M = [m_1, m_2, ..., m_n]$ be a binary mask vector of length n, where $m_i = 1$ if the corresponding sub-sentence $a_i$ is not removed, and $m_i = 0$ otherwise.
The new sentence $T_{out}$ is calculated as:
\begin{equation}
T_{out} =  M * X_A
\end{equation}

Finally, we randomly choose between $T_{mix}$ and $T_{out}$ to obtain the final augmentation results $T_{CutMixOut} $:

\begin{equation}
T_{CutMixOut}  = \begin{cases} T_{mix} & \text{with probability } p_{mix} \\
T_{out} & \text{with probability } p_{out} \end{cases}
\end{equation}
 In both CutMix and CutOut, the binary mask M is generated randomly, ensuring that different text representations can be generated from the same input.

\subsection{Generate embeddings}
For text and image backbone, our method builds upon the currently most popular visual-language pre-training (VLP) model, which has recently made significant progress and shown its rich cross-modal correspondence information and powerful visual representation capacity. Recently has made significant progress and shown its rich cross-modal correspondence information and powerful visual representation capacity the most prominent example of this is the Contrastive Language Image Pretraining (CLIP) model, which contains abundant multi-modal knowledge. This model has been used in various tasks and has produced impressive results, including video-text retrieval, image segmentation, dense prediction, and video understanding.
$768$ CLIP structure comprises two encoders - an image encoder and a text encoder, both built from a feature extractor and a projector. The feature extractor in the image encoder uses a ViT of $512$ widths, while the text encoder uses a Transformer of $512$ widths to extract features. The projectors then map these features to a 512-dimensional latent space.
Specifically, the image pre-trained model generates an embedding vector $v(I)$ that describes the contents of the image. These embeddings can then be used by mapping them to a $d$-dimensional space using a linear projection matrix $W:f(I) = W v(I)$,
where  $f(I)$ is the resulting $d$-dimensional image embedding vectors, respectively. The projection matrix $W$ can be learned from training data or pre-defined and shared across different tasks. The choice of the dimensionality $d$ of the embedding vectors depends on the pre-trained model.
To incorporate text augmentation,  each of the augmented texts uses the same text model backend and linear projection as above. We aggregate these embeddings, resulting in a final embedding vector for the original text input that captures a range of possible meanings and variations:~$f_{aug}(T) =  \sum_{i=1}^{k} W v(T_i)$,

where $T_i$ denotes the $i$th augmented text input, $v(T_i)$ epresents the embedding vector for that input, and $k$ is the total number of augmented texts generated. As previously described earlier, we use the $f_{aug}(T)$ for the downstream task of person re-identification.
After generating the separate embeddings for text and image inputs, we concatenate the text and image information as 
$J = [f_{aug}(T); f(I)]$
, where $J$ is the joint embedding vector.
The resulting concatenated embedding vector $J$ checks the similarity between two different person instances. 
The aim is to find the image and textual description pair [f(T'); f(I')] that minimizes the distance to the target image and textual description pair $(I, T)$ across all images and textual 
 descriptions captured from different cameras and sources. Mathematically, this can be expressed as
$$(I',T') = \arg \min_{I,T} d(J_{I,T}, J'_{I',T'}])$$

\section{Experiments}
\label{sec:exper}
\subsection*{Datasets}
In this paper, we used two datasets to evaluate our approach, the first one \textbf{RSTPReid} (Real Scenario Text-based Person Re-identification) \cite{RSTPReid}
 which is designed to handle real-world scenarios using the MSMT17 dataset \cite{wei2018person}. It contains 41,010 textual descriptions and 20,505 images of 4,101 individuals. Each individual has 5 images captured by 15 cameras, with each image having 2 corresponding text descriptions, and each description being at least 23 words long.
The second dataset was built based on the \textbf{PETA} \cite{PETA} dataset.
The PETA dataset comprises 19,000 images with resolutions varying from 17 x 39 to 169 x 365 pixels, each image labeled with 61 binary and 4 multi-class attributes. The binary attributes include a comprehensive list of relevant characteristics such as demographics (e.g. gender and age group), appearance (e.g. hairstyle), clothing style for the upper and lower body (e.g. casual or formal), and accessories. The four multi-class attributes include 11 basic color classifications for footwear, hair, upper-body clothing, and lower-body clothing.
We used the labeled attribute for each image to create a text description with max 21 words. For each attribute, we construct a sentence that includes the attribute and the corresponding verb. For instance, " wearing a skirt," "wearing long sleeves," "has long hair," " wearing sunglasses," and " holding a bag." 
\begin{table}[]
    \centering
    \begin{tabular}{c c c c}
        Dataset & \#Images & \#Texts&\#Cameras\\\hline
        RSTPReid & 20505& 41010&15\\\hline
        PETA & 1400 & 1400&6\\ \hline
    \end{tabular}
    \caption{Describes the dataset used in the person re-identification experiments. It includes the following information the total number of images, the number of text descriptions, and the number of cameras used to capture the images. }
    \label{tab:my_label}
\end{table}

\subsection{Evaluation Metrics and Compared Methods}
To evaluate the performance of the proposed approach we adopted the commonly used  Cumulative Matching Characteristics (CMC) curve which is a widely used evaluation metric for person re-identification methods. 

\begin{equation}
CMC(k) = \frac{1}{N} \sum_{i=1}^{N} [rank_i \leq k]
\end{equation}
where $N$ is the total number of query images, $rank_i$ is the rank of the correct match for the $i^{th}$ query and $[rank_i \leq k]$ is an indicator function that equals 1 if the rank of the correct match is less than or equal to $k$, and 0 otherwise.
To demonstrate the effectiveness of our proposed augmentation strategies method, we compare it with two existing text augmentation strategies: synonym-based PPDB and word embedding techniques. 
The synonym-based PPDB method generates augmented data by replacing words in the original text with their synonyms.
The idea behind word embeddings is to map each word to a low-dimensional vector in a continuous space. The vectors capture the meaning of the words, such that words with similar meanings are mapped to similar vectors.
Finally, we  compare the performance with  different input representations such as image-based (OSNet~\cite{Zhou_2019_ICCV}), text-based (DSSL~\cite{zhu2021dssl},
SSAN~\cite{ding2021semantically}, and  , IVT~\cite{IVT}) This enables us to determine which input representations are most effective for accurately identifying individuals across different datasets and scenarios.

\subsection{Experimental Results}
\begin{figure}[!h]
    \includegraphics[width=0.45\textwidth]{ 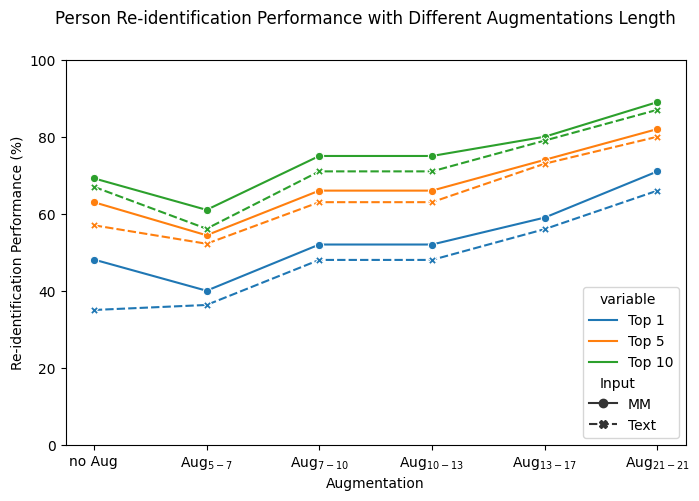}
    \caption{presents the results of using CutMixOut augmentations with varying caption lengths on person re-identification performance. The experiments were conducted on the PETA  dataset using the ViT-L14 model. }
    \label{fig:fig_aug_length}
\end{figure}

\begin{figure*}
    \includegraphics[width=0.50\linewidth]{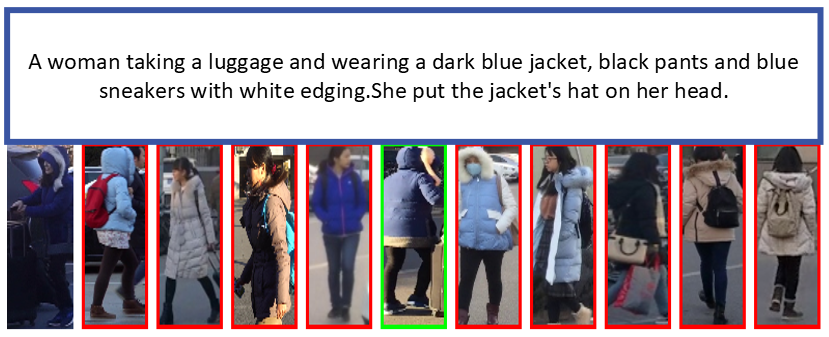} 
    \includegraphics[width=0.50\linewidth]{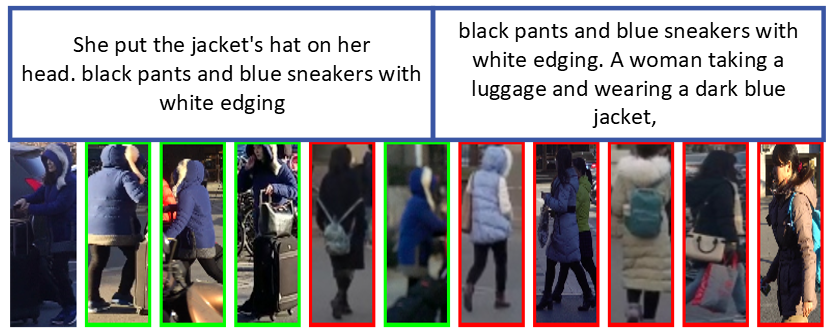} 
    \includegraphics[width=0.50\linewidth]{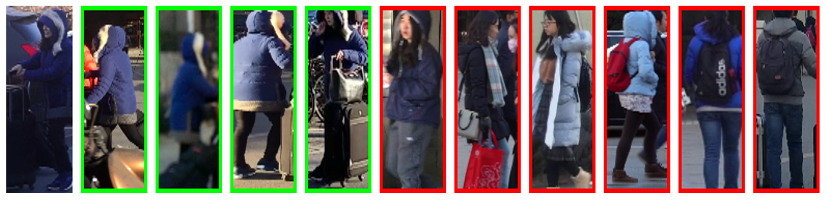}
    \includegraphics[width=0.50\linewidth]{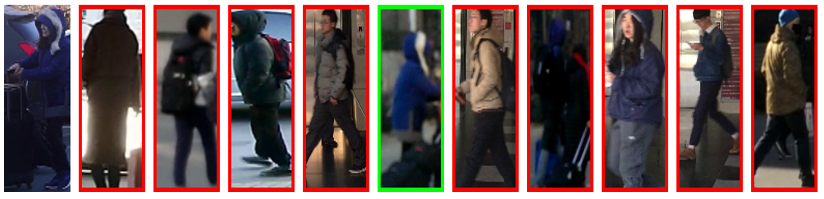}
\caption{Presents the results of the RSTPReid dataset experiment, where the top 10 ranking retrieval results of the ViT-L14 model were compared with different input types. The first row, on the left, shows the output for text-based input without augmentation, and on the right augmented text. The second row, from left to right, illustrates mixed embedding combining image and text embedding and image-based person retrieval.
}
\label{fig:abl}
\end{figure*}
The experiments are designed to achieve three main objectives. First, investigate the impact of text augmentation (TextAug) strategies during inference on improving the results. This is our proposed method. Second, compare the performance of using a mixed model or Multimodal (MM) that combines images and text with other models that use only one modality. Finally, demonstrates the superiority of the proposed augmentation strategies over existing techniques in the person re-identification problem.

The first experiment aims to evaluate the effectiveness of different input modalities for the task of person re-identification on various backbones to ensure the reliability of our findings. We assess the performance of various input types, including text, augmented text (TextAug), and Multimodal (MM) input, which combine both text and images on different architectures. The results are shown in table \ref{tab:results_1}. We used the following pre-trained models to test our approach: ViT-32, ViT-L14, ViT-B16 \cite{zhai2022lit}, and ResNet50 \cite{7780459}. Each of these models was evaluated both with and without text augmentation.
The results show that all models achieve higher top-k accuracies with text augmentation than without. The largest improvement is seen in the ViT-L14 model, which achieves a top-1 accuracy of 60.0\% with text augmentation compared to 42.8\% without. The ViT-B16 model also shows a significant improvement with text augmentation, particularly in the top-10 accuracy, where it achieves 84.5\% compared to 65.6\% without. The ResNet50 model shows less improvement with text augmentation, particularly in the top-1 accuracy, where it only improves from 39.9\% to 54.9\%.

\begin{table}
\centering
\begin{tabular}{llllll} 
\small Model &    \small TextAug     &\small Input & \small Top 1 & \small Top 5 & \small Top 10  \\ \hline 
       &            & Image  & 13.5   & 26.0    &  33.1     \\
       
\small\multirow{4}{*}{ViT 32}        &  \multirow{2}{*}{\xmark}   & \small{MM} & 20.0    & 40.2    &  48.1    \\
       &                          & \small{Text } & 31.0    & 40.2  &  45.0     \\\cmidrule{2-6} 
       & \multirow{2}{*}{\cmark} & \small{MM}  & 30.0   & 45.0    &  53.0     \\
       &                          & \small{Text}   & 35.0   & 50.0  &  57.0      \\\cmidrule{1-6} 

    &            & Image & 20.8  & 37.7    &  44.9     \\\cmidrule{2-6} 

\multirow{5}{*}{ViT-L14}       & \multirow{2}{*}{\xmark} & \small{MM} & 42.8   & 55.9    &  66.2    \\
                       &       & \small{Text}   & 40.9   & 55.2  &  61.5    \\\cmidrule{2-6} 

                         & \multirow{2}{*}{\cmark} &\small{MM}  & 60.0   & 79.5    &  84.1     \\
   &                          & \small{Text}   & 57.4   & 74.2  &  83.0     \\ \cmidrule{1-6} 

    &            & Image & 15.4 & 29.5    & 34.6   \\\cmidrule{2-6} 

\small\multirow{5}{*}{ViT-B16 } & \multirow{2}{*}{\xmark} & \small{MM} & 47.2  & 59.1   &  65.6     \\
                       &       & \small{Text}   &  44.7   & 56.1   &  60.1   \\\cmidrule{2-6} 

                         & \multirow{2}{*}{\cmark} &\small{MM}  & 59.8   &77.0    &  84.5    \\
   &                          & \small{Text}   &56.4   & 74.9  & 82.6     \\ \cmidrule{1-6} 

    &            & Image & 14.6 & 21.3    & 29.5    \\\cmidrule{2-6} 

\small\multirow{5}{*}{ResNet50 } &  \multirow{2}{*}{\xmark} & \small{MM} & 39.9   & 50.9    &  58.1   \\
                       &       & \small{Text}   &  37.1   & 48.4    &  54.6    \\\cmidrule{2-6} 

                         &  \multirow{2}{*}{\cmark} &\small{MM}  & 54.9   &75.8    &  81.3     \\
   &                          & \small{Text}   &53.8   & 74.4  & 80.0     \\

\end{tabular}
\caption{Shows the performance of ViT-32, ViT-L14, ViT-B16, and ResNet50 models on a person re-identification task with (\cmark) and without (\xmark) text augmentation conducted on the RSTPReid dataset and multi-modality.}
\label{tab:results_1}
\end{table}

The objective of the second set of experiments was to compare various augmentation strategies for person re-identification at inference time. The results are presented in table \ref{tab:result2}.
The results indicate that using data augmentation strategies can significantly improve the performance of the model. The model trained without any augmentation strategy performed poorly, with a top-1 accuracy of only 20.8\%.
Adding augmented images and texts to the input during inference improves the performance of the model, with a top-1 accuracy of 57.4\%. The proposed strategy, which builds on the cutmix and cutout with mixed input (image and augmented text), outperforms other strategies with a Top 1 accuracy of 60.0\% and Top 10 accuracy of 84.1\%. In contrast, using synonym substitution with mixed input or PPDB with text input as the augmentation strategy leads to lower accuracy across all $CMC(K)$ metrics. Similarly, using word embeddings with text input or mixed input leads to lower accuracy in top-1 and top-5 metrics compared to other strategies.

\begin{table}[]
\begin{adjustbox}{width=0.8\columnwidth,center}

    \centering

    \begin{tabular}{lllllll}
    Method     & Input & Top 1 & Top 5 & Top 10  \\ \hline
               & Image  & 20.8 & 37.7    & 44.9 \\

 \multirow{2}{*}{No Aug.} &  MM & 42.8  & 55.9   &  66.2     \\ 
                          & Text  &  40.9   & 55.2   &  61.5   \\\cmidrule{1-5} 

                           Synonym  &MM & 18.97   &33.3    & 43.6    \\ 
                        (PDDB) &Text &16.4   & 29.7  & 43.1    \\ \cmidrule{1-5} 

                         Synonym   &MM  & 15.4   &31.2   & 45.6   \\
                             (Word Embeddings) & Text&15.9   & 32.3  & 39.5    \\ \cmidrule{1-5}   

                              \multirow{2}{*}{TextAug (\textbf{Ours})} &\small{MM}  & 60.0   &79.5   &  84.1     \\
                             & Text  &57.4   & 74.2  & 83.0     \\ \cmidrule{1-5} 

    \end{tabular}
\end{adjustbox}    
    \caption{shows the results of different augmentation strategies used for person re-identification at inference time. The experiments were conducted on the RSTPReid  dataset using the ViT-L14 model.}
    \label{tab:result2}
\end{table}

Table~\ref{tab:inputRep} shows a comparison of person re-identification top-k rank performance between different input representations. Among these methods are DSSL, IVT, and SSAN which are text-based, and OSNet which is image-based. In addition, the table includes TextAug, which is the proposed method that leverages both text and images for person re-identification.
TextAug outperforms all other methods across all top-k rankings, achieving a top-1 rank of 60.0\%, a top-5 rank of 79.5\%, and a top-10 rank of 84.1\%. This suggests that the combination of text and images can result in more accurate person re-identification than relying solely on either modality alone.

\begin{table}[]
\small 
\begin{adjustbox}{width=0.8\columnwidth,center}
    \centering
    \begin{tabular}{llll}
    Method     &  Top 1 & Top 5 & Top 10   \\ \hline
    DSSL  \cite{zhu2021dssl} &  39.1  & 62.6   &  73.9     \\ 
    SSAN  \cite{ding2021semantically} &  43.5   & 67.8   & 77.15   \\ 
    IVT \cite{IVT} & 46.7 & 70.0 & 78.8 \\ 
    OSNet\cite{Zhou_2019_ICCV} &47.2& 66.3 & 75.4\\  
    TextAug{(\textbf{Ours})} & \textbf{60.0 } & \textbf{79.5 } &  \textbf{84.1}   \\\cmidrule{1-4}

    \end{tabular}
    \end{adjustbox}
    \caption{Comparison of person re-identification top-k rank performance between DSSL, SSAN, ONet, and our method.}
    \label{tab:inputRep}
\end{table}

Further experiments were conducted to investigate the impact of text augmentation with varying caption lengths on person re-identification performance on the PETA dataset.
The results are presented in figure \ref{fig:fig_aug_length}, which shows the top-1, top-5, top-10, and top-20 accuracy for each augmentation strategy. The first row of the table shows the results without any augmentation, which resulted in a top-1 accuracy of 23\%. Augmentation with mixed embeddings resulted in higher accuracy than text embeddings, with a top-1 accuracy of 48\% and 35\%, respectively.
Furthermore, the results demonstrate that increasing the number of words in the caption improved the accuracy. The highest top-1 accuracy was achieved using Aug${21-21}$ with mixed augmentation (60.0\%) and Aug${13-17}$ with text augmentation (56\%).

\begin{figure*}[!h]
 \centering
\begin{subfigure}{.38\textwidth}
  \centering
  \includegraphics[width=1\linewidth,scale=0.1]{ 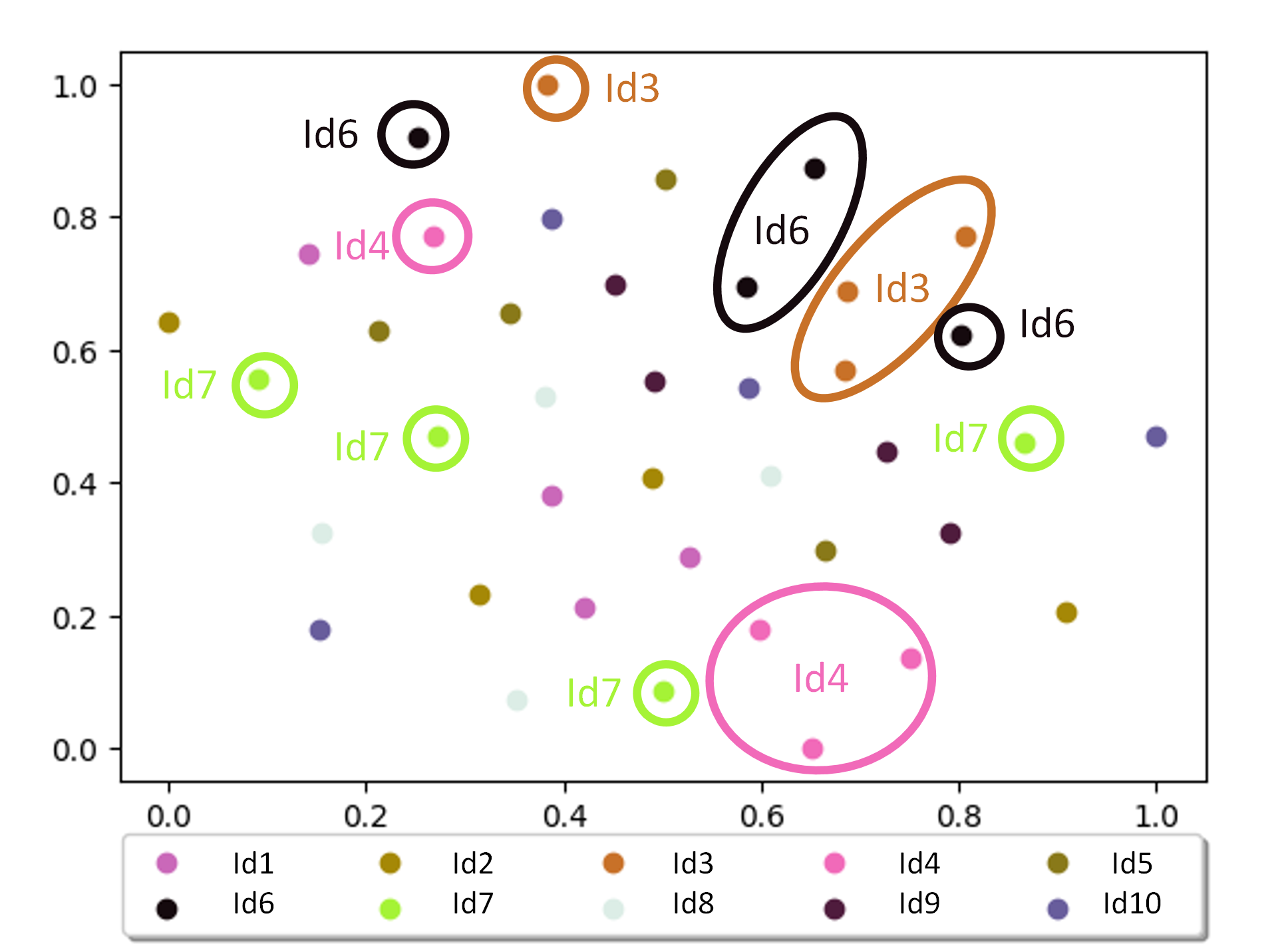} 
  \label{fig:sfig1}
\end{subfigure}%
\begin{subfigure}{.38\textwidth}
  \centering
\includegraphics[width=1\linewidth,scale=0.1]{ 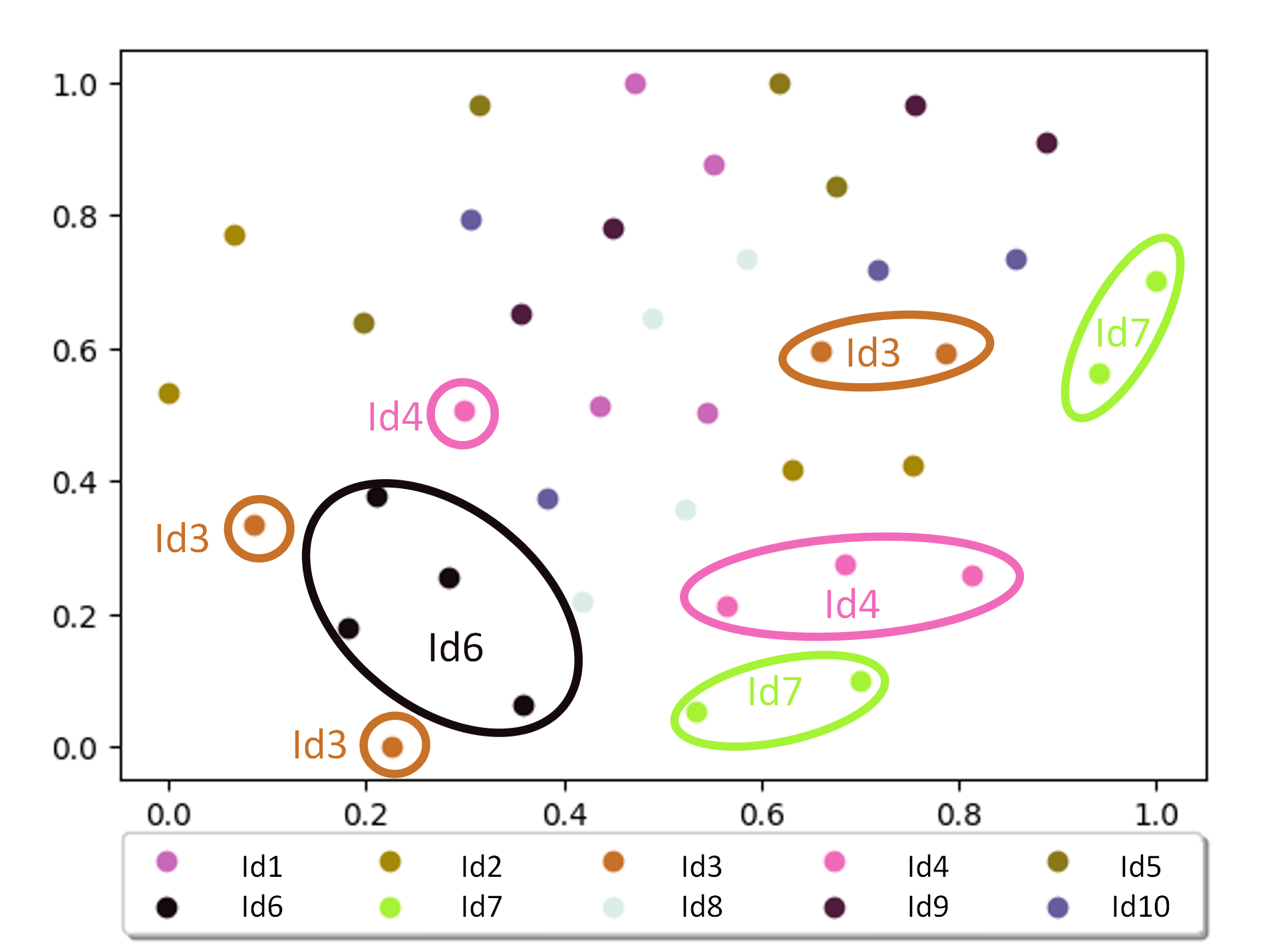}
  \label{fig:sfig2}
\end{subfigure}
\begin{subfigure}{.38\textwidth}
  \centering
\includegraphics[width=1\linewidth,scale=0.1]{ tsne_text_ViT_2_c.png}
  \label{fig:sfig2}
\end{subfigure}
\begin{subfigure}{.38\textwidth}
  \centering
\includegraphics[width=1\linewidth,scale=0.1]{ tsne_ViT_b32_mix_2_ca.png}
  \label{fig:sfig2}
\end{subfigure}
\caption{Display the t-SNE results obtained from experiments on person re-identification using the ViT-L14 model on the RSTDreid dataset. Each colored point represents a unique individual, and the figure includes the following input types: (a) image input, (b) text-based input without augmentation, (c) augmented text-based input, and (d) mixed input that combines both image and text embeddings }
\label{fig:tsne}
\end{figure*}

\subsection{Qualittive Analysis}
Figure \ref{fig:abl} provides a visual representation of the outcomes from our experimentation on person re-identification using the ViT-L14 model. The experiment aimed to investigate the impact of different input types on retrieval accuracy.
In the scenario involving image-based or text-based input without any augmentation, only one correct retrieval is observed, highlighting the inherent limitations of relying solely on either visual features or textual descriptions for person re-identification.
However, when augmented text is introduced, a notable improvement in the accuracy of text-based retrieval is observed, resulting in four correct retrievals. Additionally, in the case of Mixed Embedding (Image and Text), the results are highly effective, with the first four retrieval outcomes being accurate.
To visually analyze the distribution of individuals in the feature space and identify clusters of similar individuals for re-identification purposes, we employed T-SNE plots generated from our person re-identification study. In Figure \ref{fig:tsne}, each colored point corresponds to a distinct individual. Notably, the embeddings generated solely from the image backbone are widely scattered, lacking close clustering.
However, when text matching is employed, we observed an improvement in the results, as the embeddings with identical identities became closer together. The results were further improved when augmented text was introduced as input to the text embedding backbone, particularly when the augmented text was merged with image embeddings.

\section{Conclusion}
This study proposes TextAug, a novel approach to multimodal person re-identification by combining image and text data using text augmentation techniques at inference time. The experiments conducted demonstrate that \emph{TextAug} is an effective method for improving the performance of image-based person re-identification models, particularly for models with transformer architectures such as ViT. The proposed approach improves the generalization of deep learning models in NLP and the robustness of person re-identification models. By generating additional variations of the input text at inference time, the proposed approach achieves improved person re-identification performance and provides a comprehensive input for the distance function.

{\small
\bibliographystyle{ieee_fullname}
\bibliography{PaperForReview}

\begin{thebibliography}{10}\itemsep=-1pt

\bibitem{Chen_2018}
Dapeng Chen, Hongsheng Li, Xihui Liu, Yantao Shen, Jing Shao, Zejian Yuan, and Xiaogang Wang.
\newblock Improving deep visual representation for person re-identification by global and local image-language association.
\newblock In Vittorio Ferrari, Martial Hebert, Cristian Sminchisescu, and Yair Weiss, editors, {\em Computer Vision -- ECCV 2018}, pages 56--73, Cham, 2018. Springer International Publishing.

\bibitem{8100140}
Kan Chen, Trung Bui, Chen Fang, Zhaowen Wang, and Ram Nevatia.
\newblock Amc: Attention guided multi-modal correlation learning for image search.
\newblock In {\em 2017 IEEE Conference on Computer Vision and Pattern Recognition (CVPR)}, pages 6203--6211, 2017.

\bibitem{Dai2019}
Zuozhuo Dai, Mingqiang Chen, Xiaodong Gu, Siyu Zhu, and Ping Tan.
\newblock Batch dropblock network for person re-identification and beyond, 2018.

\bibitem{Chowdhury}
Abir Das, Anirban Chakraborty, and Amit~K. Roy-Chowdhury.
\newblock Consistent re-identification in a camera network.
\newblock In David Fleet, Tomas Pajdla, Bernt Schiele, and Tinne Tuytelaars, editors, {\em Computer Vision -- ECCV 2014}, pages 330--345, Cham, 2014. Springer International Publishing.

\bibitem{PETA}
Yubin Deng, Ping Luo, Chen~Change Loy, and Xiaoou Tang.
\newblock Pedestrian attribute recognition at far distance.
\newblock {\em Proceedings of the 22nd ACM international conference on Multimedia}, 2014.

\bibitem{cutout}
Terrance DeVries and Graham Taylor.
\newblock Improved regularization of convolutional neural networks with cutout.
\newblock 08 2017.

\bibitem{DeVries2017}
Terrance DeVries and Graham~W. Taylor.
\newblock Improved regularization of convolutional neural networks with cutout, 2017.

\bibitem{ding2021semantically}
Zefeng Ding, Changxing Ding, Zhiyin Shao, and Dacheng Tao.
\newblock Semantically self-aligned network for text-to-image part-aware person re-identification, 2021.

\bibitem{Edunov2018}
Sergey Edunov, Myle Ott, Michael Auli, and David Grangier.
\newblock Understanding back-translation at scale, 2018.

\bibitem{NIPS2013_7cce53cf}
Andrea Frome, Greg~S Corrado, Jon Shlens, Samy Bengio, Jeff Dean, Marc\textquotesingle~Aurelio Ranzato, and Tomas Mikolov.
\newblock Devise: A deep visual-semantic embedding model.
\newblock In C.J. Burges, L. Bottou, M. Welling, Z. Ghahramani, and K.Q. Weinberger, editors, {\em Advances in Neural Information Processing Systems}, volume~26. Curran Associates, Inc., 2013.

\bibitem{Guo_2020}
Hongyu Guo.
\newblock Nonlinear mixup: Out-of-manifold data augmentation for text classification.
\newblock {\em Proceedings of the AAAI Conference on Artificial Intelligence}, 34(04):4044--4051, Apr. 2020.

\bibitem{Guo2019AugmentingDW}
Hongyu Guo, Yongyi Mao, and Richong Zhang.
\newblock Augmenting data with mixup for sentence classification: An empirical study.
\newblock {\em ArXiv}, abs/1905.08941, 2019.

\bibitem{He2016}
Kaiming He, Xiangyu Zhang, Shaoqing Ren, and Jian Sun.
\newblock Deep residual learning for image recognition.
\newblock In {\em 2016 IEEE Conference on Computer Vision and Pattern Recognition (CVPR)}, pages 770--778, 2016.

\bibitem{7780459}
Kaiming He, Xiangyu Zhang, Shaoqing Ren, and Jian Sun.
\newblock Deep residual learning for image recognition.
\newblock In {\em 2016 IEEE Conference on Computer Vision and Pattern Recognition (CVPR)}, pages 770--778, 2016.

\bibitem{Jindal2020}
Amit Jindal, Dwaraknath Gnaneshwar, Ramit Sawhney, and Rajiv~Ratn Shah.
\newblock Leveraging bert with mixup for sentence classification (student abstract).
\newblock {\em Proceedings of the AAAI Conference on Artificial Intelligence}, 34(10):13829--13830, Apr. 2020.

\bibitem{https://doi.org/10.48550/arxiv.1412.2306}
Andrej Karpathy and Li Fei-Fei.
\newblock Deep visual-semantic alignments for generating image descriptions, 2014.

\bibitem{7299073}
Benjamin Klein, Guy Lev, Gil Sadeh, and Lior Wolf.
\newblock Associating neural word embeddings with deep image representations using fisher vectors.
\newblock In {\em 2015 IEEE Conference on Computer Vision and Pattern Recognition (CVPR)}, pages 4437--4446, 2015.

\bibitem{Kobayashi2018}
Sosuke Kobayashi.
\newblock Contextual augmentation: Data augmentation by words with paradigmatic relations, 2018.

\bibitem{krizhevsky2017imagenet}
Alex Krizhevsky, Ilya Sutskever, and Geoffrey~E Hinton.
\newblock Imagenet classification with deep convolutional neural networks.
\newblock {\em Communications of the ACM}, 60(6):84--90, 2017.

\bibitem{Koestinger2012}
Martin Köstinger, Martin Hirzer, Paul Wohlhart, Peter~M. Roth, and Horst Bischof.
\newblock Large scale metric learning from equivalence constraints.
\newblock In {\em 2012 IEEE Conference on Computer Vision and Pattern Recognition}, pages 2288--2295, 2012.

\bibitem{Li2017IdentityAwareTM}
Shuang Li, Tong Xiao, Hongsheng Li, Wei Yang, and Xiaogang Wang.
\newblock Identity-aware textual-visual matching with latent co-attention.
\newblock {\em 2017 IEEE International Conference on Computer Vision (ICCV)}, pages 1908--1917, 2017.

\bibitem{8100034}
Shuang Li, Tong Xiao, Hongsheng Li, Bolei Zhou, Dayu Yue, and Xiaogang Wang.
\newblock Person search with natural language description.
\newblock In {\em 2017 IEEE Conference on Computer Vision and Pattern Recognition (CVPR)}, pages 5187--5196, 2017.

\bibitem{li2021diverse}
Yulin Li, Jianfeng He, Tianzhu Zhang, Xiang Liu, Yongdong Zhang, and Feng Wu.
\newblock Diverse part discovery: Occluded person re-identification with part-aware transformer.
\newblock In {\em Proceedings of the IEEE/CVF Conference on Computer Vision and Pattern Recognition}, pages 2898--2907, 2021.

\bibitem{ma2012local}
Bingpeng Ma, Yu Su, and Fr{\'e}d{\'e}ric Jurie.
\newblock Local descriptors encoded by fisher vectors for person re-identification.
\newblock In {\em Computer Vision--ECCV 2012. Workshops and Demonstrations: Florence, Italy, October 7-13, 2012, Proceedings, Part I 12}, pages 413--422. Springer, 2012.

\bibitem{Matsukawa2016}
Tetsu Matsukawa, Takahiro Okabe, Einoshin Suzuki, and Yoichi Sato.
\newblock Hierarchical gaussian descriptor for person re-identification.
\newblock In {\em 2016 IEEE Conference on Computer Vision and Pattern Recognition (CVPR)}, pages 1363--1372, 2016.

\bibitem{mignon2012pcca}
Alexis Mignon and Fr{\'e}d{\'e}ric Jurie.
\newblock Pcca: A new approach for distance learning from sparse pairwise constraints.
\newblock In {\em 2012 IEEE conference on computer vision and pattern recognition}, pages 2666--2672. IEEE, 2012.

\bibitem{Nam2016DualAN}
Hyeonseob Nam, Jung-Woo Ha, and Jeonghee Kim.
\newblock Dual attention networks for multimodal reasoning and matching.
\newblock {\em 2017 IEEE Conference on Computer Vision and Pattern Recognition (CVPR)}, pages 2156--2164, 2016.

\bibitem{PPDB}
Ellie Pavlick, Pushpendre Rastogi, Juri Ganitkevitch, Benjamin Durme, and Chris Callison-Burch.
\newblock Ppdb 2.0: Better paraphrase ranking, fine-grained entailment relations, word embeddings, and style classification.
\newblock pages 425--430, 01 2015.

\bibitem{Qian_2020}
Jingjing Qian, Wei Jiang, Hao Luo, and Hongyan Yu.
\newblock Stripe-based and attribute-aware network: a two-branch deep model for vehicle re-identification.
\newblock {\em Measurement Science and Technology}, 31(9):095401, jun 2020.

\bibitem{Clip}
Alec Radford, Jong~Wook Kim, Chris Hallacy, Aditya Ramesh, Gabriel Goh, Sandhini Agarwal, Girish Sastry, Amanda Askell, Pamela Mishkin, Jack Clark, Gretchen Krueger, and Ilya Sutskever.
\newblock Learning transferable visual models from natural language supervision, 2021.

\bibitem{7780382}
S. Reed, Z. Akata, H. Lee, and B. Schiele.
\newblock Learning deep representations of fine-grained visual descriptions.
\newblock In {\em 2016 IEEE Conference on Computer Vision and Pattern Recognition (CVPR)}, pages 49--58, Los Alamitos, CA, USA, jun 2016. IEEE Computer Society.

\bibitem{IVT}
Xiujun Shu, Wei Wen, Haoqian Wu, Keyu Chen, Yiran Song, Ruizhi Qiao, Bo Ren, and Xiao Wang.
\newblock See finer, see more: Implicit modality alignment for text-based person retrieval.
\newblock In Leonid Karlinsky, Tomer Michaeli, and Ko Nishino, editors, {\em Computer Vision -- ECCV 2022 Workshops}, pages 624--641, Cham, 2023. Springer Nature Switzerland.

\bibitem{Simonyan2015}
Karen Simonyan and Andrew Zisserman.
\newblock Very deep convolutional networks for large-scale image recognition, 2014.

\bibitem{survey}
Nikhil~Kumar Singh, Manish Khare, and Harikrishna~B. Jethva.
\newblock A comprehensive survey on person re-identification approaches: Various aspects.
\newblock {\em Multimedia Tools Appl.}, 81(11):15747–15791, may 2022.

\bibitem{NIPS2013_2d6cc4b2}
Richard Socher, Milind Ganjoo, Christopher~D Manning, and Andrew Ng.
\newblock Zero-shot learning through cross-modal transfer.
\newblock In C.J. Burges, L. Bottou, M. Welling, Z. Ghahramani, and K.Q. Weinberger, editors, {\em Advances in Neural Information Processing Systems}, volume~26. Curran Associates, Inc., 2013.

\bibitem{Sun2018}
Ziruo Sun, Xiushan Nie, Xiaoming Xi, and Yilong Yin.
\newblock Cfvmnet: A multi-branch network for vehicle re-identification based on common field of view.
\newblock In {\em Proceedings of the 28th ACM International Conference on Multimedia}, MM '20, page 3523–3531, New York, NY, USA, 2020. Association for Computing Machinery.

\bibitem{Tan_2021}
Hongchen Tan, Xiuping Liu, Yuhao Bian, Huasheng Wang, and Baocai Yin.
\newblock Incomplete descriptor mining with elastic loss for person re-identification.
\newblock {\em {IEEE} Transactions on Circuits and Systems for Video Technology}, 32(1):160--171, jan 2022.

\bibitem{Wang_2018}
Guanshuo Wang, Yufeng Yuan, Xiong Chen, Jiwei Li, and Xi Zhou.
\newblock Learning discriminative features with multiple granularities for person re-identification.
\newblock In {\em Proceedings of the 26th {ACM} international conference on Multimedia}. {ACM}, oct 2018.

\bibitem{wang-yang-2015}
William~Yang Wang and Diyi Yang.
\newblock That{'}s so annoying!!!: A lexical and frame-semantic embedding based data augmentation approach to automatic categorization of annoying behaviors using {\#}petpeeve tweets.
\newblock In {\em Proceedings of the 2015 Conference on Empirical Methods in Natural Language Processing}, pages 2557--2563, Lisbon, Portugal, Sept. 2015. Association for Computational Linguistics.

\bibitem{wei2018person}
Longhui Wei, Shiliang Zhang, Wen Gao, and Qi Tian.
\newblock Person transfer gan to bridge domain gap for person re-identification, 2018.

\bibitem{Wieting17}
John Wieting, Jonathan Mallinson, and Kevin Gimpel.
\newblock Learning paraphrastic sentence embeddings from back-translated bitext, 2017.

\bibitem{yi2014}
Dong Yi, Zhen Lei, Shengcai Liao, and Stan~Z. Li.
\newblock Deep metric learning for person re-identification.
\newblock In {\em 2014 22nd International Conference on Pattern Recognition}, pages 34--39, 2014.

\bibitem{Yun_2019}
S. Yun, D. Han, S. Chun, S. Oh, Y. Yoo, and J. Choe.
\newblock Cutmix: Regularization strategy to train strong classifiers with localizable features.
\newblock In {\em 2019 IEEE/CVF International Conference on Computer Vision (ICCV)}, pages 6022--6031, Los Alamitos, CA, USA, nov 2019. IEEE Computer Society.

\bibitem{yun2019cutmix}
Sangdoo Yun, Dongyoon Han, Seong~Joon Oh, Sanghyuk Chun, Junsuk Choe, and Youngjoon Yoo.
\newblock Cutmix: Regularization strategy to train strong classifiers with localizable features, 2019.

\bibitem{zhai2022lit}
Xiaohua Zhai, Xiao Wang, Basil Mustafa, Andreas Steiner, Daniel Keysers, Alexander Kolesnikov, and Lucas Beyer.
\newblock Lit: Zero-shot transfer with locked-image text tuning.
\newblock {\em CVPR}, 2022.

\bibitem{Zhang2021}
Enwei Zhang, Xinyang Jiang, Hao Cheng, Ancong Wu, Fufu Yu, Ke Li, Xiaowei Guo, Feng Zheng, Wei-Shi Zheng, and Xing Sun.
\newblock One for more: Selecting generalizable samples for generalizable reid model, 2020.

\bibitem{Zhang2015}
Xiang Zhang, Junbo Zhao, and Yann LeCun.
\newblock Character-level convolutional networks for text classification.
\newblock In C. Cortes, N. Lawrence, D. Lee, M. Sugiyama, and R. Garnett, editors, {\em Advances in Neural Information Processing Systems}, volume~28. Curran Associates, Inc., 2015.

\bibitem{Zhou_2019_ICCV}
Kaiyang Zhou, Yongxin Yang, Andrea Cavallaro, and Tao Xiang.
\newblock Omni-scale feature learning for person re-identification.
\newblock In {\em Proceedings of the IEEE/CVF International Conference on Computer Vision (ICCV)}, October 2019.

\bibitem{Zhou2019}
Kaiyang Zhou, Yongxin Yang, Andrea Cavallaro, and Tao Xiang.
\newblock Omni-scale feature learning for person re-identification, 2019.

\bibitem{RSTPReid}
Aichun Zhu, Zijie Wang, Yifeng Li, Xili Wan, Jing Jin, Tian Wang, Fangqiang Hu, and Gang Hua.
\newblock Dssl: Deep surroundings-person separation learning for text-based person retrieval, 2021.

\bibitem{zhu2021dssl}
Aichun Zhu, Zijie Wang, Yifeng Li, Xili Wan, Jing Jin, Tian Wang, Fangqiang Hu, and Gang Hua.
\newblock Dssl: Deep surroundings-person separation learning for text-based person retrieval, 2021.

\end{thebibliography}
}

\end{document}